%% file: ofnoise.tex
\documentclass{article} 
\usepackage{nips14submit_e,times}
\usepackage{url}

\usepackage{graphicx}
\usepackage{amsmath,amssymb} 
\usepackage{color}
\usepackage{bbm}
\include{header}

\usepackage{xr}
\pdfoutput=1

\title{Beyond Brightness Constancy: Learning Noise Models for Optical Flow}

\author{
Dan Rosenbaum\\
School of Computer Science and Engineering\\
Hebrew University of Jerusalem\\
\url{www.cs.huji.ac.il/~danrsm} \\
\And
Yair Weiss\\
School of Computer Science and Engineering\\
Hebrew University of Jerusalem\\
\url{www.cs.huji.ac.il/~yweiss} \\
}

\nipsfinalcopy 

\begin{document}

\maketitle

\begin{abstract}
Optical flow is typically estimated by minimizing a ``data
cost'' and an optional regularizer. While there has been much work on
different regularizers many modern algorithms still use a data cost
that is not very different from the ones used over 30 years ago: a
robust version of brightness constancy or gradient constancy.
In this paper we leverage the recent availability of ground-truth
optical flow databases in order to learn a data cost. Specifically we
take a generative approach in which the data cost models the
distribution of noise after warping an image according to the flow and
we measure the ``goodness'' of a data cost by how well it matches the
true distribution of flow warp error. Consistent with current
practice, we find that robust versions of gradient constancy are
better models than simple brightness constancy but a learned GMM
that models the density of patches of warp error gives a much better fit than any existing assumption
of constancy. This significant advantage of the GMM is due to an
explicit modeling of the spatial structure of warp errors, a feature
which is missing from almost all existing data
costs in optical flow.
Finally, we show how a good density model of warp error patches can be used for optical flow estimation on whole images. We replace the data cost by the expected patch log-likelihood (EPLL), and show how this cost can be optimized iteratively using an additional step of denoising the warp error image. The results of our experiments are promising and show that patch models with higher likelihood lead to better optical flow estimation.
\end{abstract}

\section{Introduction}

Despite being a longstanding topic of study in computer vision, the current state-of-the-art optical flow estimation results are far
from being satisfactory. This is especially evident when performance
is evaluated on outdoor scenes with large occlusions and fast
motions~\cite{GeigerLU12,ButlerWSB12}. In the last two years ground truth
flow for such scenes has been made available either using synthetic
scenes~\cite{ButlerWSB12} or by accurate laser range finders that provide
flow for stationary points in the scene~\cite{GeigerLU12}. 

Like many problems in computer vision, optical flow estimation is
commonly solved by optimizing a function derived from a certain
assumed model.  The assumed model can be typically divided to a data
cost model which reflects the assumptions on the way the flow should
correspond to the images, and a regularizer that reflects the prior
assumptions on typical flow fields. 
Since the functions optimized are usually not convex, most algorithms
only achieve approximate solutions and so another critical component
in the algorithm is the optimization procedure.

In order to improve performance of flow estimation one can choose to
improve any of these three components: the regularizer, the data cost
and the  optimizer. While much recent work has explored using
different regularizers (e.g.~\cite{ZimmerBW11,sun2012layered})
or different optimizers (e.g.~\cite{xu2012scale,sun2014quantitative,yamaguchi2013robust}) there has
been relatively little work on the data term. A notable exception is
the recent work of Vogel and Roth~\cite{vogel2013evaluation} which
compares the effect of  different versions of brightness or gradient
constancy on the performance of optical flow algorithms.

Brightness constancy and gradient constancy are in a sense
``hand-crafted'' data costs. Is it possible  to leverage the availability of ground truth
flow datasets in order to {\em learn} a data cost for optical flow?
A step in this direction was taken by Sun et al~\cite{sun2008learning} who
used a Fields of Experts distribution over the error term and learned
$3 \times 3$ filters that defined the data cost. The learned cost was
similar to gradient constancy but with irregular filters.

\begin{figure}
\centering
\renewcommand{\tabcolsep}{5pt}
\begin{tabular}{cc}
\begin{tabular}{cc}
first image $I_1$ & ground-truth flow $v$\\
\includegraphics[width=0.3\textwidth]{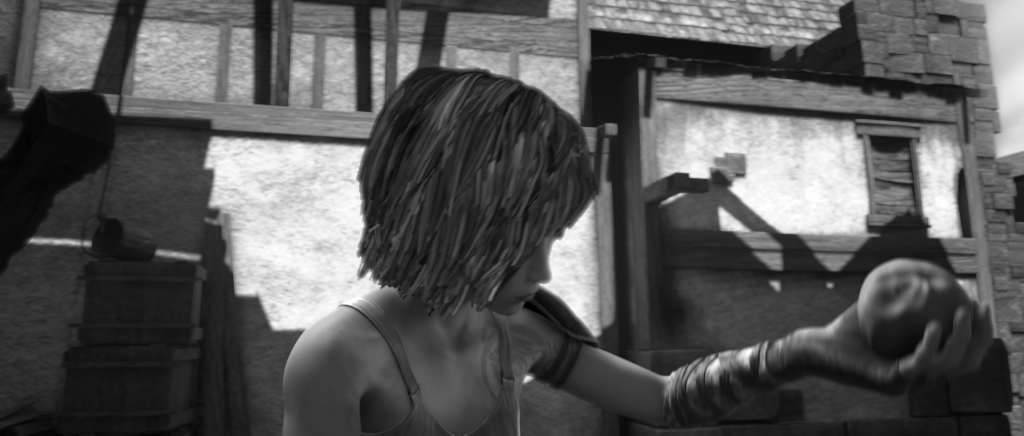} &
\includegraphics[width=0.3\textwidth]{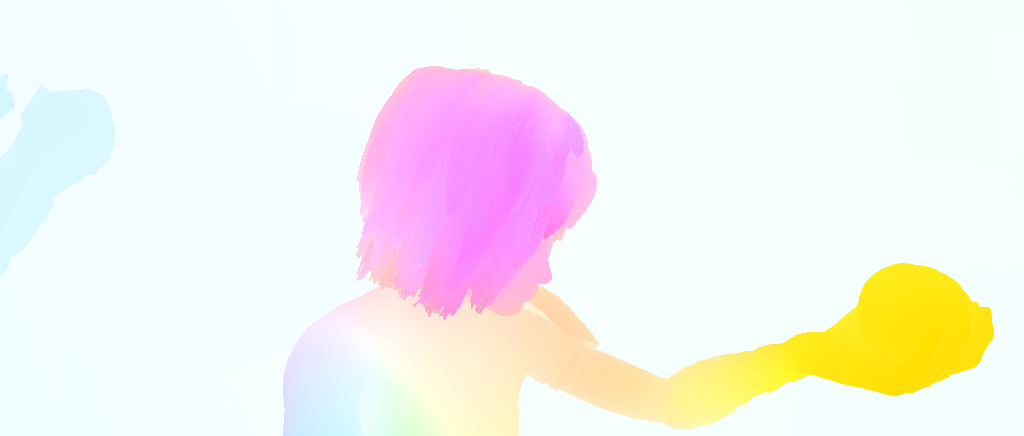} \\
second image $I_2$ & flow warp error $D_v$ \\
\includegraphics[width=0.3\textwidth]{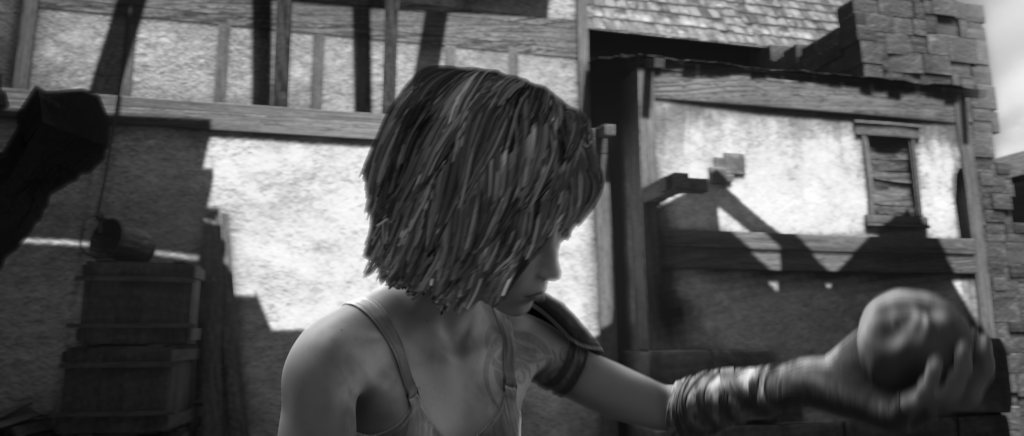} &
\includegraphics[width=0.3\textwidth]{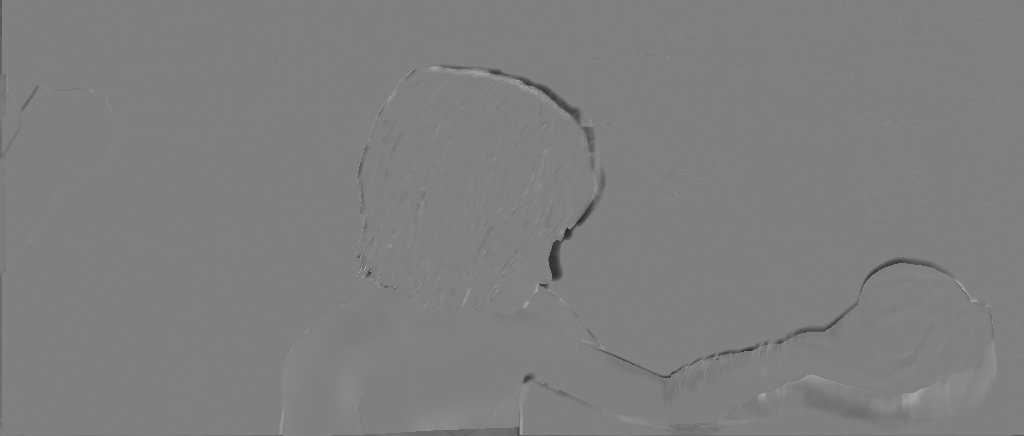}
\end{tabular} &
\begin{tabular}{c}
patches of $D_v$\\
\includegraphics[width=0.18\textwidth]{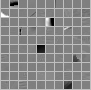}
\end{tabular}
\end{tabular}

\caption{Flow warp error. The second image is warped backwards according to the ground-truth flow and subtracted from the first image. The resulting warp error image is then divided to small patches. In this paper we use a database of such patches in order to learn a data term for optical flow. In contrast to models used in common algorithms, the warp error has an evident spatial structure and is far from being isotropic noise.}
\label{fig:warpError}
\end{figure}

In this paper we take a generative approach in which the data cost
models the distribution of noise after warping an image according to
the flow. Under the ideal brightness constancy assumption, when we
backwards-warp the second image according to the optical flow we
should obtain an image that is identical to the first one
(figure~\ref{fig:warpError}). In real images, of course, we never get exact matches and we call the difference between the warped second image and the first image the ``flow warp error.'' Different data costs for optical flow give different penalties for this flow warp error. 

Here we measure the ``goodness'' of a data cost by how well it matches the
true distribution of flow warp error. By focusing on patches of flow
warp errors we 
can measure this ``goodness'' robustly and efficiently.  Consistent with current
practice, we find that robust versions of gradient constancy are
better models than simple brightness constancy but a learned Gaussian Mixture Model (GMM)
density model of the error gives a much better fit than any existing
assumption of constancy. This significant advantage of the GMM is due to an
explicit modeling of the spatial structure of warp errors, a feature
which to the best of our knowledge is missing from the vast majority of existing data costs in optical flow.

A second question we address here, is how a patch model of flow warp error can be used for flow estimation of whole images. To do so, we replace the image data cost by the \emph{expected patch log-likelihood} (EPLL) term  introduced by \cite{zoran2011learning}. We also propose a method for optimizing this cost, which is based on \emph{half-quadratic splitting}~\cite{wang2008new,krishnan2009fast}. Our method boils down to an iterative algorithm consisting of two steps. In the first step we solve a flow estimation problem with a simple brightness constancy cost, and in the second step we ``denoise'' the resulting image of flow warp error using a patch density model. The results of our experiments are promising and show that patch models with higher likelihood lead to better optical flow estimation.

\subsection{Optical flow data costs} \label{sec:ofDataCosts}

\subsubsection*{Brightness Constancy} 
Perhaps the most common data cost simply penalizes the gray-scale distance between every pixel in the first image $I_1$ and its corresponding location in the second image $I_2$ according to the flow $v$. 
This is equivalent to creating a warped image $I_2^v$ by warping back $I_2$ according to $v$ and then subtracting the warped image from $I_1$.
In classic optical flow algorithms like Horn and Schunck~\cite{horn1981determining} and Lucas-Kanade~\cite{lucas1981iterative} the squared distance is summed over all pixels (SSD) resulting in:
\begin{equation} \label{eq:J_BCL2}
J_{BCL2} = \sum_p (I_1(p) - I_2^v(p))^2 ~.
\end{equation} 
In later extensions like Black and Anandan~\cite{BlackA93}, more robust functions are used, e.g. the sum of absolute distances (SAD),
\begin{equation} \label{eq:J_BCL1}
J_{BCL1} = \sum_p |I_1(p) - I_2^v(p)| ~.
\end{equation}

\subsubsection*{Gradient Constancy} 
A second approach is to measure the distance of the image spatial derivatives rather than gray-scale values~\cite{brox2004high}, thus allowing a constant change in gray-scale. Denoting by $I_{1x}$,$I_{1y}$ and $I_{2x}$,$I_{2y}$ the horizontal and vertical derivatives of the first and second image, this is equivalent to warping $I_{2x}$ and $I_{2y}$ according to $v$ and subtracting them from $I_{1x}$ and $I_{1y}$,
\begin{equation} \label{eq:J_GCL2}
J_{GCL2} = \sum_p (I_{1x}(p) - I_{2x}^v(p))^2 + \sum_p (I_{1y}(p) - I_{2y}^v(p))^2 ~.
\end{equation}
Once again, the quadratic function can be replaced by a more robust function like the absolute value,
\begin{equation} \label{eq:J_GCL1}
J_{GCL1} = \sum_p |I_{1x}(p) - I_{2x}^v(p)| + \sum_p |I_{1y}(p) - I_{2y}^v(p)| ~.
\end{equation}

\subsubsection*{Census} 
An increasingly popular approach to deal with smooth changes of gray-scale between images is to replace the gray-scale by some monotone ranking in a certain neighborhood. In the Census transform~\cite{zabih1994non}, the data cost at a pixel $p$ counts the number of neighboring pixels $q$ that change their sign relative to p, 
\begin{equation} \label{eq:J_CEN}
J_{CEN} = \sum_p \sum_q \mathbbm{1}_{\left[sign(I_1(q) - I_1(p)) \neq sign(I_2^v(q) - I_2^v(p))\right]} ~,
\end{equation}

A convex approximation of the Census transform can be formulated by replacing the indicator and sign functions by the absolute value, resulting in the centralized sum of absolute distance (CSAD) data cost~\cite{vogel2013evaluation}
\begin{equation} \label{eq:J_CSAD}
J_{CSAD} = \sum_p \sum_q | (I_1(q) - I_1(p)) - (I_2^v(q) - I_2^v(p)) | ~.
\end{equation}

One drawback of all the above costs is that they are all sums of local costs and lack the modeling of spatial structure.
Figure \ref{fig:warpError} shows the warp error $D_v = I_1-I_2^v$ of images from the Sintel dataset~\cite{ButlerWSB12}, using the provided ground-truth flow. The warp error images show an evident spatial structure. Even when looking at small random patches from the dataset, the structure is clearly observed. In particular patches tend to be flat and close to zero but occasionally contain an edge in some orientation. 


\section{The data cost as a noise model}
Using a generative approach to flow estimation from a pair of images $I_1$ and $I_2$, 
it can be assumed that the first image $I_1$ is generated as 
\begin{equation} \label{eq:generativeModel}
  I_1 = I_2^v + w
\end{equation}
where $w$ is a random noise image generated from some density model. 
In this view, different data costs that are functions of the warp error $D_v = I_1-I_2^v$, are equivalent to different density models of $w$:
\begin{equation} \label{eq:pexpj}
Pr(I_1|I_2;v) = Pr(D_v) = \frac{1}{Z}e^{-\lambda J(D_v)} 
\end{equation}
Notice that according to equation \ref{eq:generativeModel}, the warp error $D_v$ is equal to the noise $w$ and thus equation \ref{eq:pexpj} is also a density model over the additive noise $w$.

The data costs we consider above: brightness constancy (BC), gradient constancy (GC) and centralized sum of absolute differences (CSAD) are all functions of the warp error. In particular, they can all be expressed as the $l_2$-norm or $l_1$-norm of a linear transformation of $D_v$. Therefore we can formulate them as density models as follows:

\textbf{Brightness Constancy L2} Exponentiating $J_{BCL2}$ (equation \ref{eq:J_BCL2}), we obtain a multidimensional Gaussian which is a product of independent Gaussians with  variance $1/2\lambda$.
\begin{equation} \label{eq:P_BCL2}
  Pr(D_v) = \frac{1}{Z} e^{-\lambda \sum_p D_v(p)^2} = \frac{1}{Z}e^{-\lambda ||d_v||_2^2} 
\end{equation} 
where $d_v$ is a vector created by concatenating all pixels in $D_v$.

\textbf{Brightness Constancy L1} Exponentiating $J_{BCL1}$ (equation \ref{eq:J_BCL1}), we obtain a multidimensional  Laplace distribution which is a product of independent Laplacians  with variance $1/2\lambda$.
\begin{equation} \label{eq:P_BCL1}
  Pr(D_v) = \frac{1}{Z} e^{-\lambda \sum_p |D_v(p)|} = \frac{1}{Z}e^{-\lambda ||d_v||_1} 
\end{equation} 

\textbf{Gradient Constancy L2} Exponentiating $J_{GCL2}$ (equation \ref{eq:J_GCL2}), we obtain a multidimensional Gaussian with inverse  covariance matrix $\lambda A^\top A$ where $A$ is a derivative matrix that computes the horizontal and vertical derivatives at each pixel. Since this matrix is not invertible we add $\epsilon I$. 
\begin{equation} \label{eq:P_GCL2}
  Pr(D_v) = \frac{1}{Z} e^{-\lambda \sum_p D_{vx}(p)^2 + D_{vy}(p)^2} \approx \frac{1}{Z}e^{- d_v^\top (\lambda A^\top A + \epsilon I) d_v} 
\end{equation} 

\textbf{Gradient Constancy L1} Exponentiating $J_{GCL1}$ (equation \ref{eq:J_GCL1}), we obtain a multidimensional Laplace distribution. As in GCL2, we add $\epsilon I$ to make this distibution normalizable, and since the normalization constant $Z$ cannot be found in closed form we use Hamiltonian Annealed Importance Sampling to approximate it~\cite{souldick2011hais}.
\begin{equation} \label{eq:P_GCL1}
  Pr(D_v) = \frac{1}{Z} e^{-\lambda \sum_p |D_{vx}(p)| + |D_{vy}(p)|} \approx \frac{1}{Z}e^{- ||(\lambda A + \epsilon I) d_v||_1} 
\end{equation} 

\textbf{Centralized Sum of Absolute Differences} Exponentiating $J_{CSAD}$ using a $5 \times 5$ neighborhood around each pixel $p$ (equation \ref{eq:J_GCL1}),  we obtain a multidimensional Laplace distribution. Now the derivative matrix $A$ contains more rows corresponding to all the differences between $p$ and each pixel $q$ in the $5 \times 5$ neighborhood. Like in GCL1 we need to add $\epsilon I$ and  approximate the normalization constant using Hamiltonian Annealed Importance Sampling. 
\begin{equation} \label{eq:P_CSAD}
  Pr(D_v) = \frac{1}{Z} e^{-\lambda \sum_p \sum_q |D_v(q)-D_v(p)|} \approx \frac{1}{Z}e^{- ||(\lambda A_{5 \times 5} + \epsilon I) d_v||_1} 
\end{equation} 

\subsection{Comparing different data costs}

Perhaps the most direct way of comparing different data costs is by
evaluating the relative performance of optical flow algorithms that
use these costs. This is the approach taken in~\cite{vogel2013evaluation,sun2008learning}. The main drawback of this approach is that the flow predicted by an algorithm is usually the result of a complicated, nonconvex optimization and many parameters can influence the final result. For example, Sun et al~\cite{sun2014quantitative} reported that changing the number of levels in the pyramid used for coarse to fine optimization can dramatically change the performance of some algorithms on the Sintel benchmark. 

Here we take an alternative approach. We consider the data costs as density models  on $D_v$, 
and ask: which of these density models best fits the distribution of actual patches of flow warp errors?
The primary method we use to estimate the goodness of fit is the {\em
  average log likelihood on held out data}. It is well known that this log
likelihood can be equivalently written as a constant minus the KL
divergence between the empirical distribution and the density model. Thus the model that gives highest log likelihood to held out data is also the model whose distribution is most similar to the empirical distribution. 

We create a dataset of flow warp error $D_v$ using
the Sintel dataset. First we use the ground-truth flow to warp the images backwards, then we subtract the warped images from their corresponding preceding images, and finally we divide the resulting images to $8 \times 8$ patches. 
Following~\cite{rosenbaum2013learning} we divide the training set of Sintel into two parts:  $708$ image pairs in training and $333$ pairs were used for testing. All model parameters (e.g. $\lambda,\epsilon$ discussed above) were learned on the training set using maximum likelihood.  
We then compare the likelihood of different density models on a random sample of patches from the test set.
We repeat this process for each of the three passes of Sintel: \emph{albedo}, \emph{clean} and \emph{final}, resulting in three separate training sets and test sets.
Since all our results are very similar on all the three passes we focus here only on the \emph{final} pass.

\begin{figure}
\centering
\includegraphics[width=0.8\textwidth]{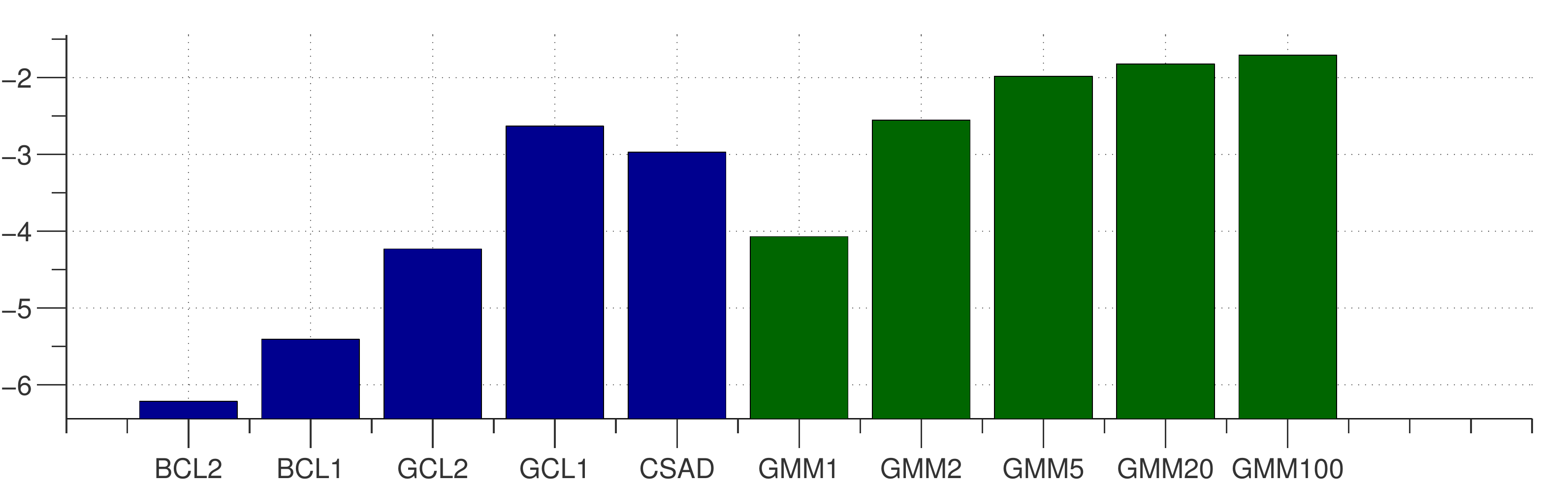}
\caption{Log-likelihood on held-out data from Sintel's final pass. The learned GMM noise models (green) are compared to the common data cost noise models (blue). Consistent with common practice, gradient constancy better fits the data than brightness constancy and robust (L1) costs are better than Gaussian (L2). However, a GMM with 100 components outperforms all other models.}
\label{fig:modelsLLH}
\end{figure}

The resulting likelihood for the above models are shown in figure \ref{fig:modelsLLH} (in blue). 
The main things to note are that the $l_1$-norm is better than the $l_2$-norm, that the constant gradient assumption is better than the constant brightness assumption, and that the convex approximation of the census transform is very similar to the gradient constancy assumption. 
These findings agree with the comparison of optical flow estimation using different data costs reported by \cite{vogel2013evaluation}.

\begin{figure}
\centering
\renewcommand{\tabcolsep}{1.5pt}
\begin{tabular}{cc}
\begin{tabular}{c}
GT\\\includegraphics[width=0.155\textwidth]{figs/patches_final.png}
\end{tabular} & 
\begin{tabular}{ccccc}
BCL2 & BCL1 & GCL2 & GCL1 & CSAD \\
\includegraphics[width=0.155\textwidth]{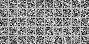} &
\includegraphics[width=0.155\textwidth]{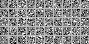} &
\includegraphics[width=0.155\textwidth]{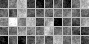} &
\includegraphics[width=0.155\textwidth]{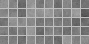} &
\includegraphics[width=0.155\textwidth]{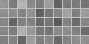} \\[0.3cm]
GMM1 & GMM2 & GMM5 & GMM20 & GMM100 \\
\includegraphics[width=0.155\textwidth]{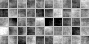} &
\includegraphics[width=0.155\textwidth]{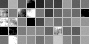} &
\includegraphics[width=0.155\textwidth]{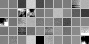} &
\includegraphics[width=0.155\textwidth]{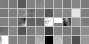} &
\includegraphics[width=0.155\textwidth]{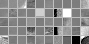} 
\end{tabular}
\end{tabular}
\caption{Patch samples of warp error from Sintel's final pass (GT) and randomly generated using the different noise models. Top: samples generated using the common data cost models. Bottom: samples generated using the learned GMM models. The patches generated from GMM100 demonstrate a very similar structure to the ground-truth patches.}
\label{fig:modelsSamples}
\end{figure}

Another way to measure how well models capture the true statistics is by {\em generating samples}.   Patches created from a certain model, typically satisfy the underlying assumptions of the model. Therefore, a visual resemblance to the ground-truth suggests that the patches were generated from a better model. We can see in figure \ref{fig:modelsSamples} (top row), that the patches generated from GCL1 and CSAD are the most similar to the ground-truth patches. Although those patches seem to model the flatness correctly, evidently, they fail to model the occasional structure that is present in the ground-truth.

\section{Learning the noise model}
Following the recent success of learning Gaussian mixture models (GMM) in natural image statistics~\cite{zoran2012natural} and as prior models for optical flow~\cite{rosenbaum2013learning}, we use the training set to estimate GMMs with a different number of components.
Every component of the GMM is a multivariate Gaussian with zero mean and a full covariance matrix.
We train the GMM using the Expectation Maximization (EM) algorithm on mini-batches from the training set.
It is important to note that the GMM has far more parameters than the common data cost models and thus we emphasize that the 
models are tested on the held-out test set to assure no overfitting occurs.

The resulting likelihood of all models is shown in figure \ref{fig:modelsLLH}. 
The results show that: 1. a single Gaussian model (GMM1) has a very similar likelihood to the L2 Gradient Constancy model (GCL2); 
2. a GMM with 2 components (GMM2) is similar to the robust L1 Gradient Constancy model (GCL1); and  
3. a GMM model with 100 components (GMM100) outperforms all other models.

We also use the learned GMMs to generate random patches and compare them to the ground-truth patches and to random patches generated by the common data cost models. Figure \ref{fig:modelsSamples} shows that the patches generated by GMM100 resemble the ground-truth patches more than other models do.

\subsection{What does the GMM learn?}
\begin{figure}
\centering
\renewcommand{\tabcolsep}{1.5pt}
\begin{tabular}{cllc}
 &  & \small{leading eigenvectors} & \small{randomly generated samples} \\
\begin{tabular}{c} \tiny{GMM1}\end{tabular} & \begin{tabular}{c}\tiny{$\pi=1$}\end{tabular} & 
\begin{tabular}{c} \includegraphics[width=0.13333\textwidth]{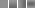} \end{tabular}&
\begin{tabular}{c} \includegraphics[width=0.5\textwidth]{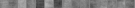} \end{tabular}\\
\begin{tabular}{c} \tiny{GMM2}\end{tabular} & \begin{tabular}{c} \tiny{$\pi=0.72$} \\ \tiny{$\pi=0.28$} \end{tabular} & 
\begin{tabular}{c} \includegraphics[width=0.25\textwidth]{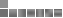} \end{tabular}&
\begin{tabular}{c} \includegraphics[width=0.5\textwidth]{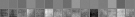} \end{tabular}\\

\begin{tabular}{c} \tiny{GMM100}\end{tabular} & \begin{tabular}{c} \tiny{$\pi=0.41$} \\ \tiny{$\pi=0.04$} \end{tabular} & 
\begin{tabular}{c} \includegraphics[width=0.067\textwidth]{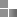} \end{tabular}&
\begin{tabular}{c} \includegraphics[width=0.5\textwidth]{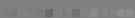} \end{tabular}\\

 & \begin{tabular}{c} \tiny{$\pi \approx 10^{-4}$}\end{tabular} & 
\begin{tabular}{c} \includegraphics[width=0.1\textwidth]{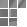} \end{tabular}&
\begin{tabular}{c} \includegraphics[width=0.5\textwidth]{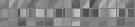} \end{tabular}

\end{tabular}
\caption{The leading eigenvectors and randomly generated samples for the components of GMM1 and GMM2, and for some selected components of GMM100. GMM2 captures outliers by modeling flat patches and noisy patches separately. GMM100 explicitly distinguishes between flat patches with different variance, and patches with different types of edges.}
\label{fig:GMMcomp}
\end{figure}

We investigate the learned GMM models to understand what makes them better than the common data cost models, 
In figure \ref{fig:GMMcomp} we show the components of GMM1 (which contains only one component), GMM2 and some selected components of GMM100. Each component is shown in a different row. Each of those components is a Gaussian, and to illustrate its preference,  we show the leading eigenvectors of the covariance matrix corresponding to $95\%$ of the cumulative eigenvalues (i.e. corresponding to $95\%$ of the variance) re-organized as patches. In addition, we show in each row a set of patches that were randomly generated using the corresponding Gaussian. 

For the single Gaussian model (GMM1) which essentially estimates the covariance of the patches, we can see that the leading eigenvectors of the covariance correspond to smooth changes in patches. This is also seen in the randomly generated patches. This behavior is similar to what the gradient constancy with $l_2$ norm (GCL2) models. 

For GMM2, the figure shows that the first component favors flat patches in a much stronger manner than GMM1. 
This can be seen both in the generated samples and by the fact that $95\%$ of the variance is expressed by the single flat eigenvector. 
In contrast, the second component of GMM2 allows the patches to be much more noisy than GMM1, and needs more eigenvectors to reach $95\%$ of the variance. 
Similarly to the robustness characteristic of GCL1, the behavior of GMM2, can be viewed as a form of outlier detection where  $72\%$ of the errors are essentially just an additive constant and $28\%$ of them are allowed to be very noisy.  

For GMM100,  we show only a few selected components ordered by decreasing mixing weights. The first components, capture the flatness assumption, and each component allows a random constant change with a different variance. Looking at components with lower mixing weights we see components that capture more interesting structure. {\em Most components are dedicated to edges in certain orientations and shifts.} Intuitively this model learns that most of the time the warped patch and the true patch will differ by an additive constant, but when this is not the case, the difference is not simply white noise. {\em Rather this ``noise'' is extremely structured and is well approximated locally by an oriented edge.} In retrospect, this assumption is very intuitive and is related to the process of occlusion. Differences between the original patches and warped patches that are not simple additive constants are most commonly the result of occlusion and disocclusion. Since the occluded objects have spatial structure, so does the warp error. While this assumption is very intuitive, we are not aware of any optical flow data cost that utilizes it.

\section{Optical Flow Estimation}

We now show how a density model of warp error patches can be used for optical flow estimation.
A common way to estimate optical flow from a pair of images is by minimizing an energy function containing a data cost depending on the input images and a regularizer on the flow field $R(f)$. Using our generative assumption (equations \ref{eq:generativeModel},\ref{eq:pexpj}), the data cost we wish to minimize is equal to the log density model of the warp error of the whole image $Pr(D_v)$. 

Given a patch density model, one way to define the image density model, introduced in \cite{zoran2011learning}, is to measure the \emph{expected patch log-likelihood} (EPLL) in the image.
Recall that $d_v$ is a vector representation of the warp error image, and denote by $P_i$ a matrix that extracts the $i$'th patch from it. The EPLL cost can be written as:
\begin{equation} \label{eq:costEPLL}
J(v) =  - \sum_i \log Pr(P_i ~ d_v) + \lambda R(v)
\end{equation}
The exact minimization of the cost defined in equation \ref{eq:costEPLL} is not tractable.
The first reason is that the warp error $d_v$ is a non-convex function of the flow $v$. The common way to overcome this is by iteratively approximating $d_v$ as a linear function of the flow (by taking the Taylor expansion of the image intensities around the current warp). A second problem is that even after the linearization of $d_v$ the density model might cause the minimization to be intractable. To solve this for any density model we use the method of \emph{half-quadratic splitting} as presented in \cite{wang2008new,krishnan2009fast}, combined with the EPLL image denoising method of \cite{zoran2011learning}. 
In \emph{half-quadratic splitting}, we introduce a new variable $r$, resulting in the following new cost:
\begin{equation} \label{eq:costSplit}
  J(v,r) =  - \sum_i \log Pr(P_i ~ r) + \beta ||d_v-r||_2^2 + \lambda R(v)
\end{equation}
This cost is approximately minimized by alternatingly solving for $v$ and for $r$ and by gradually increasing $\beta$. Note that once $\beta$ is big enough, $r$ is forced to be close to $d_v$ and we return to the original EPLL cost (equation \ref{eq:costEPLL}). We next describe the 2 steps performed in each iteration:

\textbf{r-step:} When $v$ is fixed, the third term in equation \ref{eq:costSplit} is constant and solving for $r$ is equivalent to the problem of image denoising using a prior on clean patches. The ``noisy'' image in this case is the warp error image $d_v$, and the cost function on the difference $d_v-r$ is equivalent to the assumption that the noise model is Gaussian, isotropic and with variance $1/\beta$. Solving for $r$ can be done using the EPLL denoising algorithm introduced in \cite{zoran2011learning}, where any patch model can be used (assuming that a patch denoising method is provided). 

\textbf{v-step:} When $r$ is fixed, solving for $v$ is equivalent to estimating the optical flow using a simple brightness constancy data cost on the image, where the first image is ``fixed'' according to $r$. To see this recall that $d_v = I_1-I_2^v$ and so defining a new image $I^r_1=I_1-r$ results in the cost: $\argmin_v ||I^r_1-I_2^v||_2^2  + \frac{\lambda}{\beta} R(v)$.

\subsection{Experiments}
\begin{figure}
\centering
\includegraphics[width=0.8\textwidth]{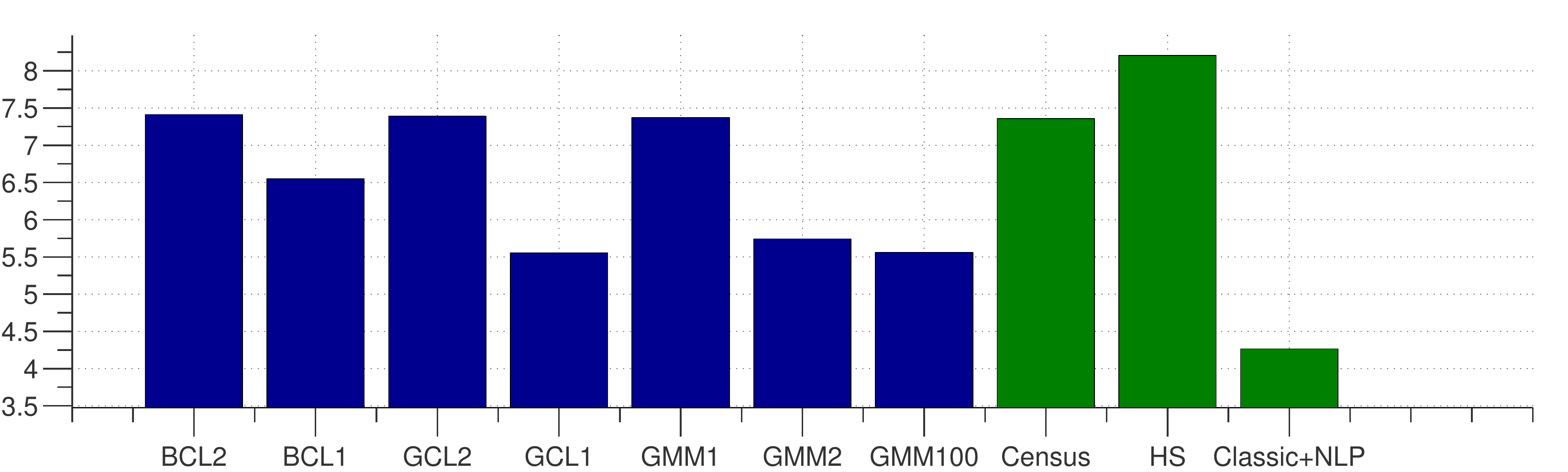}
\caption{Average end-point-error of the optical flow estimated on 20 Sintel images. Blue: Our proposed EPLL method using different warp error patch models. Models with higher likelihood generally lead to lower error. Green: Reference algorithms. Even with a fairly simple flow regularizer, our EPLL method, combined with a good warp error patch model, is able to reach an error that is comparable to some of the top performing algorithms.}
\label{fig:sintel20aepe}
\end{figure}

To test the method proposed above, we use it to estimate the optical flow in the Sintel dataset using different warp error patch models. The estimation is performed in a coarse-to-fine manner such that in every level we run 20 iterations, each consisting of one r-step and one v-step. During the 20 iterations we gradually increase $\beta$ to assure the original cost (equation \ref{eq:costEPLL}) is decreasing. We use a common regularizer that penalizes the spatial derivatives of the flow using the $l_1$ norm~\cite{vogel2013evaluation,sun2014quantitative}, and optimize it in each v-step using the \emph{iteratively reweighted least squares} (IRLS) method. In each v-step we perform one image warp and linearization. The r-step is performed using the EPLL denoising software published in \cite{zoran2011learning}. We start the process using an initial flow estimate in the coarsest level that was computed using a standard gradient constancy algorithm.

For reference, we also compare the EPLL method to other algorithms, which we run on the same 20 Sintel images. We use the software provided by \cite{sun2014quantitative} to run their implementation of Horn and Schunck (HS) and Classic+NLP which is one of the top performing algorithms in the Sintel and KITTI datasets. We also use the software by \cite{vogel2013evaluation} which implements the Census transform data cost and is also one of the top performing algorithms in Sintel and KITTI. For all those algorithms we use the default parameters as suggested in their software kits.

The results are shown in figure \ref{fig:sintel20aepe}. It can be seen that the performance of the EPLL with different warp error models is correlated to the likelihood of the models as shown in figure \ref{fig:modelsLLH}. In general, models with higher likelihood lead to flow estimation with smaller average end-point-error.  The results also show that our EPLL method, combined with a good warp error patch model, is able to estimate the optical flow with error that is comparable to the reference algorithms: with a good warp error model EPLL outperforms the classic Horn and Schunck algorithm and the Vogel et al. implementation that also uses an L1 regularizer. The Classic+NLP algorithm uses a stronger, nonlocal regularizer and outperforms all the methods that use an L1 regualrizer in these experiments.

While we have found that all other things being equal, better warp noise models lead to better optical flow performance, our experiments indicate that the optimization method and the regularizers can be just as important. In particular, we find that the result of our ``v-step'' which uses a standard coarse-to-fine optimization procedure is often suboptimal and gives higher cost than the ground truth flow. This suggests that more powerful optimization methods are needed.

\section{Discussion}
In this paper we use a generative approach to evaluate and learn optical flow data costs. 
By focusing on patches of flow warp errors we measure the likelihood of different models robustly and efficiently. We show that evaluating the likelihood of
existing data costs, largely agrees with common practice. We find that a learned GMM gives a better fit to the true distribution and show that it is related to the seperate representation of flat patches and different edge orientations.
This intuitive structure that mirrors the spatial structure of occluding objects in natural scenes, has not been used in existing data costs for optical flow.
Finally, we show how good patch models of warp error can lead to better performance in flow estimation. We define a new data cost which models the expected patch log likelihood and propose a method to optimize it. The results of our experiments show that using models with higher likelihood leads to better estimation. Even though we use a fairly simple flow regularizer, our EPLL method, combined with a good warp error patch model, is able to estimate the optical flow with error that is comparable to some of the top performing algorithms. We are confident that further research on improving the optimization, and combining our novel data cost with a strong regularizer, can lead to improved optical flow estimation.

\bibliographystyle{plain}
\bibliography{bib}

\end{document}

%% file: header.tex
\usepackage{graphicx,subfigure}
\usepackage{epsfig}
\usepackage{amsmath}
\usepackage{amssymb}
\usepackage{algorithm}
\usepackage{algorithmic}
\usepackage{url}
\usepackage{enumerate}
\usepackage{amsfonts}
\usepackage{boxedminipage}
\usepackage{xcolor}
\algsetup{indent=2em}

\newcommand{\BlackBox}{\rule{1.5ex}{1.5ex}}

\DeclareMathOperator*{\argmin}{argmin} 

\renewcommand{\eqref}[1]{Equation~(\ref{#1})}

\newcommand{\coursename}{(67577) INTRODUCTION TO MACHINE LEARNING}

\newcommand{\handout}[5]{
   \renewcommand{\thepage}{#1-\arabic{page}}
   \noindent
   \begin{center}
   \framebox{
      \vbox{
    \hbox to 5.78in { {\bf \coursename}
         \hfill #2 }
       \vspace{4mm}
       \hbox to 5.78in { {\Large \hfill #5  \hfill} }
       \vspace{2mm}
       \hbox to 5.78in { {\it #3 \hfill #4} }
      }
   }
   \end{center}
   \vspace*{4mm}
}

